%% file: main.tex
\documentclass[10pt,twocolumn,letterpaper]{article}

\usepackage[pagenumbers]{iccv} % To force page numbers, e.g. for an arXiv version


\definecolor{iccvblue}{rgb}{0.21,0.49,0.74}
\usepackage[pagebackref,breaklinks,colorlinks,allcolors=iccvblue]{hyperref}

\usepackage{multirow}
\usepackage{comment}
\usepackage{bbding}

\usepackage[accsupp]{axessibility}  % Improves PDF readability for those with disabilities.

\newlength\savewidth

\def\paperID{2461} % *** Enter the Paper ID here
\def\confName{ICCV}
\def\confYear{2025}

\title{Beyond Walking: A Large-Scale Image-Text Benchmark\\ for Text-based Person Anomaly Search}

\author{Shuyu Yang$^1$ \quad 
Yaxiong Wang$^{2}\footnotemark[1]$ \quad 
Li Zhu$^{1}\footnotemark[1]$ \quad 
Zhedong Zheng$^3\footnotemark[1]$ \\
$^1$Xi'an Jiaotong University \quad 
$^2$Hefei University of Technology \quad 
$^3$University of Macau\\
{\tt\small \{ysy653, wangyx15\}@stu.xjtu.edu.cn, 
zhuli@mail.xjtu.edu.cn, 
zhedongzheng@um.edu.mo}
}

\begin{document}
\input{sec/0_abstract}

\input{sec/1_intro}

\input{sec/2_related}

\input{sec/3_bench}

\input{sec/4_method}

\input{sec/5_experiment}

\input{sec/6_conclusion}

{
    \small
    \bibliographystyle{ieeenat_fullname}
    \bibliography{main}
}

\appendix
\input{sec/X_suppl}

% {
%     \small
%     \bibliographystyle{ieeenat_fullname}
%     \bibliography{main}
% }

\end{document}

%% file: sec/0_abstract.tex
\twocolumn[{
\renewcommand\twocolumn[1][]{#1}
\maketitle
\begin{center}
\setlength{\abovecaptionskip}{8pt} 
\setlength{\belowcaptionskip}{0pt} 
\centering
\vspace{-.3in}
\includegraphics[height=2.2in, width=6.7in]{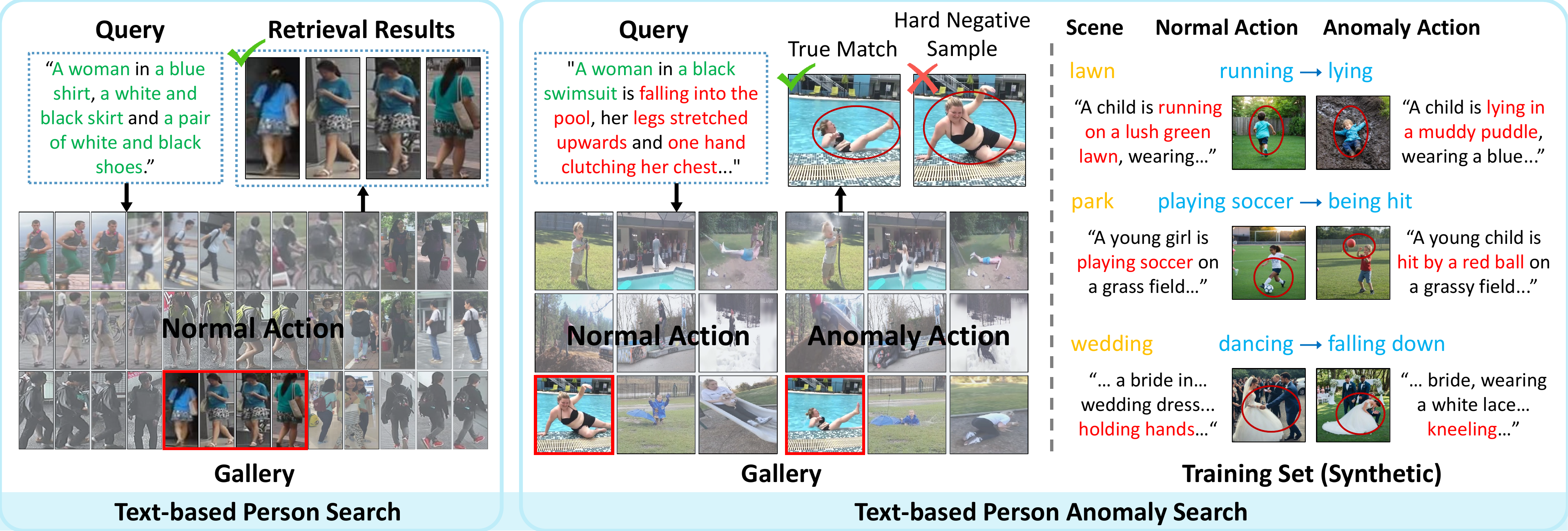}
\vspace{-.05in}
\captionof{figure}
{\textbf{Comparison of our proposed task, \ie, Text-based Person Anomaly Search (\emph{right}) \emph{vs.} Traditional Text-Based Person Search (\emph{left}).} Traditional text-based person search primarily focuses on the appearance of individuals and often overlooks action information, if any. In contrast, given the appearance and action description, text-based person anomaly search aims to locate the pedestrian of interest engaged in either normal or abnormal actions from a large pool of candidates. Text-based person anomaly search emphasizes the identification of pedestrian abnormal behaviors, aligning closely with real-world emergency and safety requirements.
}
\label{fig:example}
\end{center}}]

% \twocolumn[{
% \renewcommand\twocolumn[1][]{#1}
% \maketitle
% \vspace{-18pt}
% \begin{center}
%     \captionsetup{type=figure}
%     % \vspace{-.1in}
%     \includegraphics[width=\textwidth]{img/example.pdf}
%     \\
%     \captionof{figure}{\textbf{Comparison of our proposed task, \ie, Text-based Person Anomaly Search (\emph{right}) \emph{vs.} Traditional Text-Based Person Search (\emph{left}).} Traditional text-based person search primarily focuses on the appearance of individuals and often overlooks action information, if any. In contrast, given the appearance and action description, text-based person anomaly search aims to locate the pedestrian of interest engaged in either normal or abnormal actions from a large pool of candidates. Text-based person anomaly search emphasizes the identification of pedestrian abnormal behaviors, aligning closely with real-world emergency and safety requirements.
%     }
%     \label{fig:example}
% \end{center}
% }]

\footnotetext[1]{Corresponding author. The dataset, model, and code are available at \url{https://github.com/Shuyu-XJTU/CMP}.}

\begin{abstract}
Text-based person search aims to retrieve specific individuals across camera networks using natural language descriptions.
However, current benchmarks often exhibit biases towards common actions like walking or standing, neglecting the critical need for identifying abnormal behaviors in real-world scenarios. 
To meet such demands, we propose a new task, text-based person anomaly search, locating pedestrians engaged in both routine or anomalous activities via text. 
To enable the training and evaluation of this new task, we construct a large-scale image-text Pedestrian Anomaly Behavior (PAB) benchmark, featuring a broad spectrum of actions, \eg, running, performing, playing soccer, and the corresponding anomalies, \eg, lying, being hit, and falling of the same identity. 
The training set of PAB comprises $1,013,605$ synthesized image-text pairs of both normalities and anomalies, while the test set includes $1,978$ real-world image-text pairs. 
To validate the potential of PAB, we introduce a cross-modal pose-aware framework, which integrates human pose patterns with identity-based hard negative pair sampling.  Extensive experiments on the proposed benchmark show that synthetic training data facilitates the fine-grained behavior retrieval, and the proposed pose-aware method arrives at $84.93\%$ recall@1 accuracy, surpassing other competitive methods. 
%The dataset, model, and code are available at https://github.com/Shuyu-XJTU/CMP.
%We will release the dataset, code, and checkpoints to facilitate further research and ensure the reproducibility.
\end{abstract}

%% file: sec/1_intro.tex
\section{Introduction}
\label{sec:intro}

\begin{table*}[t]
  \centering
  \vspace{-.2in}
  % \small
  \resizebox{0.96\textwidth}{!}{
  \footnotesize
  \begin{tabular}{@{}lcccccccc@{}}
      \toprule
      \multirow{1}{*}{Datasets} & Modality & \multirow{1}{*}{Annotation} & \multirow{1}{*}{Content} & \#Frames  & \multirow{1}{*}{\#Texts} & \multirow{1}{*}{\#Action Types} & \multirow{1}{*}{Anomaly:Normal} & \multirow{1}{*}{Data source} \\
      \midrule
      CUHK-PEDES~\cite{li2017person} & Image, Text & Frame-level Text & Appearance & 40,206 & 80,440 & - & 0 & Collection \\ %Market~\cite{Zheng_2015_ICCV}, \etc
      ICFG-PEDES~\cite{ding2021semantically} & Image, Text & Frame-level Text & Appearance & 54,522 & 54,522 & - & 0 & Collection \\ %MSMT-17~\cite{wei2018person} \\
      RSTPReid~\cite{zhu2021dssl} & Image, Text & Frame-level Text & Appearance & 20,505 & 41,010 & - & 0 & Collection \\ %MSMT-17~\cite{wei2018person}\\
      %\multirow{2}{*}{MALS~\cite{yang2023towards}} & \multirow{2}{*}{Image \& Text} & \multirow{2}{*}{1,510,330} & \multirow{2}{*}{1,510,330} & \multirow{2}{*}{27.0} & \multirow{2}{*}{$531\times 208$} & \multirow{2}{*}{Nor.} & Appearance \& & \multirow{2}{*}{Synthesis} \\
      %&  &  &  &  &  &  & Attribute & \\
      UBnormal~\cite{acsintoae2022ubnormal} & Video & Frame-level Tag & Binary Label & 236,902 & - & 22 Anomaly & 2:3  & Synthesis \\
      ShanghaiTech~\cite{luo2017revisit} & Video & Frame-level Tag & Binary Label & 317,398 & - & 11 Anomaly & 1:18 & Collection \\
      UCF-Crime~\cite{sultani2018real} & Video & Video-level Tag & Binary Label & 13,741,393 & - & 13 Anomaly & $\ll$1:1 & Collection \\
      UCA~\cite{yuan2024towards} & Video, Text & Video-level Text & Action & 13,741,393 & 23,542 & 13 Anomaly & $\ll$1:1 & Collection \\
      \hline
      \multirow{2}{*}{{PAB} (Ours)} & \multirow{2}{*}{Image, Text} & \multirow{2}{*}{Frame-level Text} & {\bf Appearance,} & \multirow{2}{*}{{1,015,583}} &\multirow{2}{*}{{\textbf{1,015,583}}} & \multirow{1}{*}{\bf 1,600 Anomaly} & \multirow{2}{*}{{\bf{3:2}}} & Synthesis \& \\
         &  &   & {\bf Action, Scene} &  &  & {\bf 1,000 Normal} &  & Collection \\
    \bottomrule
  \end{tabular}
  }
      \vspace{-.1in}
  \caption{{\bf Dataset Characteristic Comparison.} 
  We present a comprehensive comparison between our proposed Pedestrian Anomaly Behavior (PAB) benchmark and existing text-based pedestrian search and video anomaly detection datasets in terms of data quality and quantity. Our dataset addresses the long-tail distribution challenge by incorporating a higher number of anomaly cases, while offering frame-level annotations (appearance, action, and scene) and fine-grained action types.
  }
  \label{tab:comparison}
   \vspace{-.15in}
\end{table*}

Text-based person search~\cite{li2017person,zheng20242} focuses on retrieving specific individuals from large-scale image databases using natural language descriptions. This capability is particularly valuable for user-interactive applications in smart cities, security systems, and personalized services, where image queries are often unavailable or impractical to obtain. However, current benchmarks for text-based person search suffer from significant limitations in behavioral diversity, primarily due to their bias towards common pedestrian actions. Existing datasets, including CUHK-PEDES~\cite{li2017person}, ICFG-PEDES~\cite{ding2021semantically}, and RSTReid~\cite{zhu2021dssl}, predominantly feature routine activities such as walking and standing, failing to represent the full spectrum of real-world behaviors adequately (see Fig.~\ref{fig:example}). This inherent bias significantly restricts the generalizability and practical applicability of models trained on these datasets, particularly in critical scenarios requiring anomaly behavior detection.

Moreover, current anomaly detection methodologies focus on identifying and classifying predefined events, while existing video datasets face three major limitations:  (1) limited data volume, typically containing less than 500k frames~\cite{li2013anomaly,lu2013abnormal,mehran2009abnormal}, (2) coarse annotation granularity, often restricted to binary labels (anomaly \emph{vs.} normal)~\cite{luo2017revisit,sultani2018real,wang2025UniAD}, and (3) behavior long-tail distribution, where anomaly key frames are significantly underrepresented compared to normal behaviors~\cite{adam2008robust,ramachandra2020street,yuan2024towards} (see Table~\ref{tab:comparison}). These limitations hinder the development of robust models capable of handling diverse real-world scenarios.
While UBnormal~\cite{acsintoae2022ubnormal} addresses multi-class annotations using simulated 3D environments, the gap between virtual scenes and real-world footage limits generalization in practical deployments. 
Moreover, real-world applications often demand the ability to search for specific behaviors rather than broad, predefined categories, further highlighting the need for more nuanced datasets.

To address the limitations in both fields, we introduce a new task, \ie, text-based person anomaly search. \textbf{(1)} This task extends the traditional scope of text-based person search by requiring the identification of pedestrians involved in both routine and anomalous activities. While existing methods could successfully locate a person walking, they often fail to identify the same person lying on the ground or being hit. \textbf{(2)} Comparing with the conventional anomaly detection, the proposed task further explores a fine-grained anomaly understanding beyond the binary tags,  which is critical for security events tracing and locating, emergency response, and many other security applications.
To support this task, we propose the Pedestrian Anomaly Behavior (PAB) benchmark. As summarized in Table~\ref{tab:comparison}, the PAB benchmark consists of $1,015,583$ image-text pairs, each annotated with detailed textual descriptions of the target pedestrian’s appearance, actions, and surrounding scene.  The dataset comprehensively spans 1,000 distinct normal action, such as running, performing, and playing soccer, along with 1,600 anomalies like lying, being hit, and falling. This broad coverage ensures diversity in both routine and rare pedestrian activities, while the intentionally elevated anomaly ratio (3:2) enhances its utility for training robust anomaly detection models.
To validate the potential of the PAB benchmark, we introduce a Cross-Modal Pose-aware (CMP) framework that integrates human pose patterns with identity-based hard negative pair sampling. This framework leverages the rich pose information to distinguish between normal and anomalous behaviors. Extensive experiments show that synthetic training data facilitate the person anomaly search on the real-world test set. The proposed framework also achieves a substantial improvement. In a summary, our primary contributions are:
\begin{itemize}
    \item   We pioneer the task of text-based person anomaly search that \textbf{unifies detection of both routine and anomalous activities} through natural language queries. To address the critical absence of fine-grained anomaly data, we construct the Pedestrian Anomaly Behavior (PAB) benchmark, featuring $1.01M$  image-text pairs spanning $1,600$ anomaly and $1,000$ normal action types.
    \item Departing from conventional identity-centric person search methods, we propose a Cross-Modal Pose-aware (CMP) framework that integrates text, image, and human pose patterns for representation learning. We also introduce identity-based hard negative mining by perturbing the action to distinguishing subtle behavioral differences.
   \item  Extensive experiments show that (1) synthetic training data of the proposed PAB facilitates fine-grained behavior retrieval in the real-world test set; (2) our full CMP model achieves 84.93\% recall@1 accuracy on PAB and 55.23\% recall@1 accuracy on the out-of-distribution (OOD) UCC testing, outperforming multiple competitive approaches.
\end{itemize}

%% file: sec/2_related.tex
\section{Related Work}
\noindent\textbf{Text-based Person Retrieval. }
Text-based pedestrian retrieval incorporates textual queries into large-scale pedestrian retrieval tasks, breaking the limitations of image-based~\cite{zheng2015scalable,zheng2017unlabeled,martinel2019aggregating} or attribute-based queries~\cite{lin2019improving, han2018attribute, ling2019improving}. 
Text-based pedestrian retrieval is generally more challenging than image-based pedestrian retrieval~\cite{almazan2018re,zheng2015scalable,reidyaxiong1,zheng2019joint} and general cross-modal retrieval tasks~\cite{yan2023urban, yao2023capenrich, duan2023match4match, guo2024query,pfan,pfan++} due to its cross-modal and fine-grained nature. 
The key to this task is to learn the alignment of the image and text~\cite{chen2022tipcb, niu2020improving, li2022learning, gao2021contextual,tip2017,sigir24,sigir25}.
Roughly, prevailing alignment strategies can be divided into cross-modal attention-based~\cite{li2017person, wang2022look, shu2023see, shao2022learning, park2024plot} and cross-modal non-attention~\cite{ding2021semantically, zheng2020dual, wang2022caibc, chen2022tipcb} methods. 
With the development of vision-language pre-training models~\cite{radford2021learning, li2021align}, recent works~\cite{han2021text, shu2023see, yan2022clip, jiang2023cross, bai2023rasa} have improved the robustness of learned features by transferring knowledge from large-scale generic image-text pairs.  
Inspired by the large language and multimodal models, recent studies attempt to utilize the large model to facilitate this problem~\cite{yang2023towards,tan2024harnessing}.
Yang \etal~\cite{yang2023towards} presents a large-scale synthetic image-text dataset MALS for pre-training, while Tan \etal~\cite{tan2024harnessing} generate a large-scale dataset to study transferable text-to-image ReID. % with Multimodal Large Language Models (MLLMs)
In this paper, we introduce text-based person anomaly search, focusing on fine-grained behavior retrieval.

% \vspace{-.01in}
\noindent\textbf{Person Anomaly Detection.}
Existing work on pedestrian anomaly detection typically refers to Video Anomaly Detection (VAD)~\cite{cao2024context}. The goal of VAD is to detect events in videos that deviate from normal patterns, and it has been studied under various settings: as a one-class classification 
problem~\cite{feng2021convolutional,flaborea2023multimodal,hirschorn2023normalizing,reiss2022attribute,zaheer2022generative} where training only involves normal data; as an unsupervised learning task~\cite{zaheer2022generative} where anomalies exist in training set but it is unknown which training videos contain them; and as a supervised or weakly supervised problem~\cite{zaheer2022generative, acsintoae2022ubnormal, sultani2018real} where training labels indicate anomalous video frames or videos containing anomalies.
Most of these works focus on the one-class classification problem. Consequently, many video datasets are limited in terms of the number of videos or the realism of the scenarios. 
Examples include the UCSD Ped1 and Ped2~\cite{li2013anomaly}, Avenue~\cite{lu2013abnormal}, Subway~\cite{adam2008robust}, ShanghaiTech Campus~\cite{luo2017revisit}, and UBnormal~\cite{acsintoae2022ubnormal}.
Sultani \etal ~\cite{sultani2018real} construct a realistic video dataset called UCF-Crime, which annotates $13$ pre-defined categories of video anomalies but still falls short in handling complex real-world scenarios.
Despite efforts~\cite{yuan2024towards,wu2024toward} to annotate UCF-Crime videos with text, datasets such as UCA~\cite{yuan2024towards} remain constrained (see \Cref{tab:comparison}). 
Therefore, we propose a large-scale multi-modal Pedestrian Anomaly Behavior (PAB) dataset that contains large-scale diverse normal and anomalous image-text pairs, capable of addressing more complex multi-modal anomaly behavior retrieval.

%% file: sec/3_bench.tex
\section{Person Anomaly Benchmark}
\label{sec:bench}

\begin{figure*}[t]
  \centering
   \vspace{-0.2in}
   \includegraphics[width=0.96\linewidth]{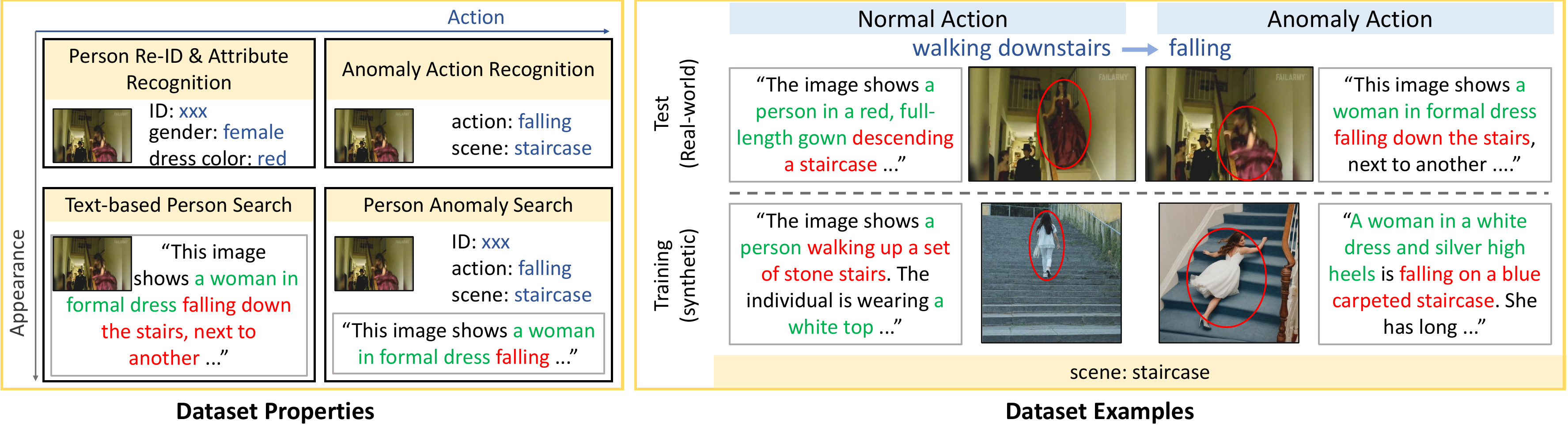}
   \vspace{-0.1in}
   \caption{\textbf{Dataset Properties (\emph{left})}. 
   Compared with existing datasets for person re-ID, attribute recognition, text-based person search, and anomaly action recognition, ours contain more detailed action and appearance descriptions for text-based anomaly search.
   \textbf{Dataset Examples (\emph{right})}.
   Our training set is synthesized, while the test set is collected from real-world videos. 
   We provide similar training samples in terms of normal and anomaly action, facilitating effective comparative analysis during model training.}
   \label{fig:datasets}
   \vspace{-.15in}
\end{figure*}

\subsection{Real-world Test Data Collection}
To establish a practical evaluation benchmark, we construct our test set using real-world videos from OOPS!~\cite{epstein2020oops}. 
% Note that this serves as a preliminary raw collection. 
% Next, we will illustrate how to perform rigorous filtering and annotation refinement to ensure data quality and task alignment.

\noindent\textbf{Anomaly and Anomaly Image Extraction.}
OOPS! videos come with timestamps indicating when an anomaly or unintentional action begins. This means that the content before the timestamp depicts normal behavior, while the content after the timestamp shows anomaly behavior. 
We extract middle frames from the video segments before and after the timestamps as pedestrian normal and anomaly images, respectively, and still
%The extracted normal and anomaly image pairs
face the following issues:
\begin{itemize}
\item Noise images, \ie, images that do not contain people or where people occupy a small proportion of the image area. To address this issue, we deploy OpenPose~\cite{cao2017realtime}
to detect human key points and eliminate undesired images.
\item  Duplicated images or images with subtle discrepancy. To overcome such noise, we apply ResNet-50~\cite{he2016deep} to extract features from the images and calculate the cosine similarity between normal and anomaly images, filtering out pairs with a similarity greater than $0.95$.
\item  {False-positive} anomaly images, \ie, behaviors in anomaly images are normal. For this problem, we conduct manual verification by three professionals with advanced education in computer science, retaining only those image pairs of certain normal and anomaly behaviors. 
Through the aforementioned steps, we obtain $989$ high-quality image pairs ($1,978$ pedestrian images), which are designated as {1:1} normal and anomaly image pairs for the test set.
\end{itemize}

\noindent\textbf{Caption Generation and Quality Control.}
Each OOPS! video is annotated with two types of captions, one ($C_n$) for normal moment and another ($C_a$) for anomaly occurring.
Directly adopting these captions as image captions does not work well, because most of the captions are short and without detailed descriptions of appearance. 
Drawing inspiration from the significant progress in the Multi-modal Large Language Model (MLLM)~\cite{bai2023qwen,chen2023shikra}, we automatically generate captions for each image, eliminating the need for costly manual annotation. Particularly, we choose Qwen2-VL~\cite{bai2023qwen} as Image Captioner.
The specific {\bf Instruction} is as follows:
``Provide a simple description of the image content within 50 words, including the appearance, attire, and actions of the main figures in the picture. Do not imagine any contents that are not in the image. Do not describe the atmosphere of the image."
The captions generated by MLLM can generally provide accurate descriptions of the appearance and actions of people in the images, including detailed information. 
However, due to inherent limitations of the model, the captions may contain minor errors. 
Therefore, we conduct {\bf Human Quality Control} on the test set. Specifically, we enlist three professionals with advanced degrees in computer science to manually correct the text data. This measure ensures that the texts accurately match the image content especially for specific normal and anomaly behaviors.

\subsection{Large-scale Training Data Generation}
To bypass the scarcity of anomaly data, we resort to utilizing the {generative model, \eg, }diffusion model, to synthesize our training data with the following steps:

\noindent{\bf Pedestrian-focused Image Generation.}
To ensure the generated images are realistic and stylistically consistent with the test set, we use captions from the OOPS! dataset. 
The $C_n$ and $C_a$ captions are input into the Realistic Vision V4.0 model~\cite{RealVisXLV4.0} to generate high-quality pedestrian images. 
It is worth noting that, before image generation, we filter the captions to ensure they describe human subjects. For each caption, we extract the subject and verify if it refers to a person, such as ``people", ``he", ``man", \etc. This process yields $6,739$ $C_n$ and $6,979$ $C_a$ captions. Additionally, for cases where both $C_n$ and $C_a$ are retained, we concatenate them to form a new $C_{a+}$. The resulting $6,669$ $C_{a+}$ captions serve as prompts for generating pedestrian images with anomaly behaviors, enhancing diversity. 
Despite the ability of the Realistic Vision V4.0 model to produce photorealistic images, it occasionally generates images with unreasonable human structures. 
To mitigate this issue, we use OpenPose~\cite{cao2017realtime} for human key point detection and eliminate images with structural errors. After generating 75 images per caption and applying image filtering, we obtain 1,013,605 high-quality, diverse pedestrian training images, with an anomalous to normal behavior ratio of approximately 3:2.

\begin{figure*}[t]
\vspace{-.2in}
  \centering
   \includegraphics[width=0.95\linewidth]{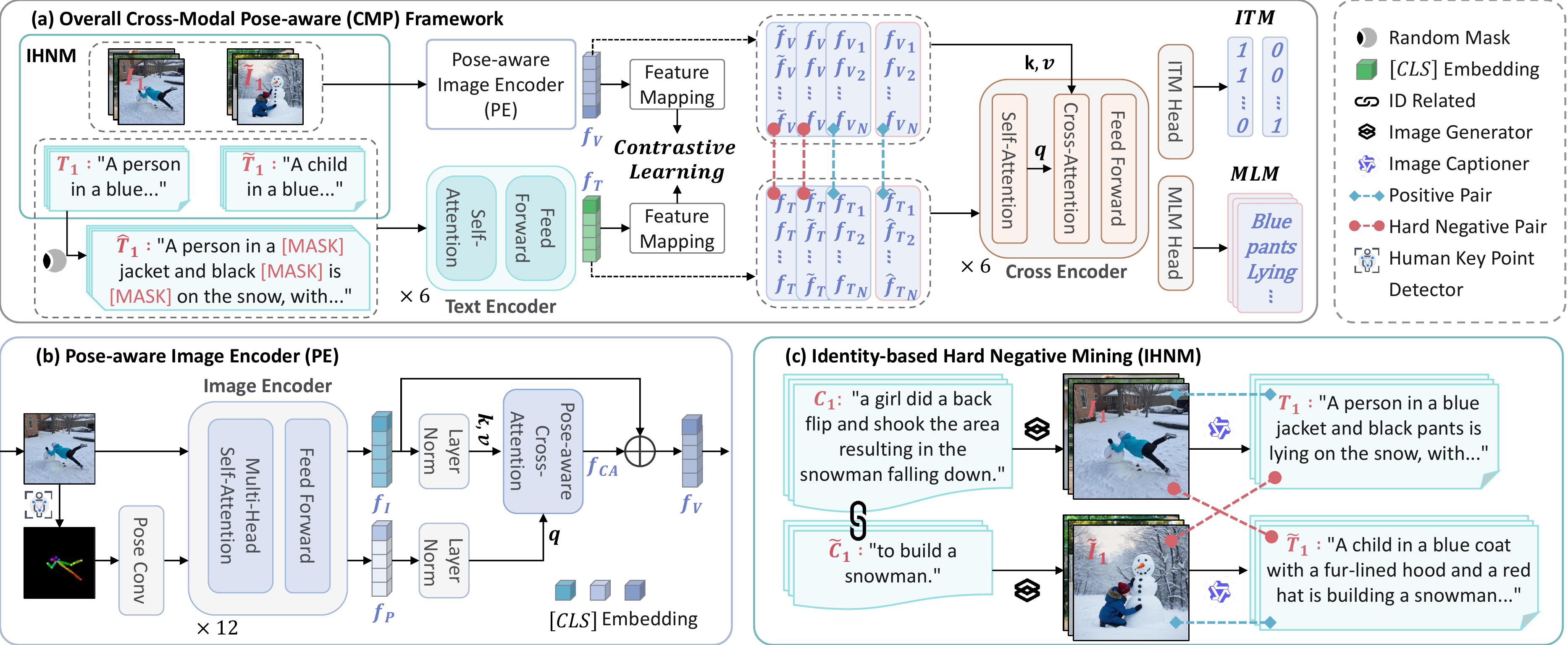}
   \vspace{-.1in}
   \caption{\textbf{(a) Overview of our Cross-Modal Pose-aware (CMP) framework}, composed of (b) a pose-aware image encoder, a text encoder, and a cross encoder. We apply (c) Identity-based Hard Negative Mining to form challenging negative pairs during training, followed by feature extraction, contrastive learning, and processing by the cross encoder with Image-Text Matching (ITM) and Mask Language Modeling (MLM) heads. The final step computes a cross-modal loss including ITM and MLM components.
   }
   \label{fig:framework}
   \vspace{-.1in}
\end{figure*}

\noindent{\bf Text Diversifying via Re-captioning.}
Similar to the caption generation process for test set image, we re-caption each synthetic image via Qwen2-VL~\cite{bai2023qwen}.
The minor errors contained in the generated caption can be acceptable for the training set.
So we directly use the generated caption as a matched text description for each training image.

\noindent{\bf Attribute Annotation.}
Aiming at pedestrian anomaly retrieval, we further enrich our dataset with annotations of actions, anomalies, and scenes. 
Given the cost of manually annotating large-scale data, we opt to leverage the multi-modal understanding capability of MLLM (Qwen2-VL~\cite{bai2023qwen}) to automatically obtain the action types, anomalous behaviors, and scenes for each image-text pair. Specially, for an anomaly image-text pair $(I, T)$, we get its aforementioned attributes by querying the MLLM with carefully crafted instructions in \emph{Suppl}. %\footnote{will be presented in our supplementary file}
Note that the attribute recognition is not the focus of this paper, we provide these attributes to support future tasks like action or scene classification.

\subsection{Dataset Analysis}
Following the steps above, we have successfully created the Pedestrian Anomaly Behavior (PAB) dataset, a large-scale, richly annotated collection. 
Its properties and examples are depicted in Fig. \ref{fig:datasets}. We compare PAB with other prominent Text-based Person Re-Identification datasets (\ie, CUHK-PEDES~\cite{li2017person}, ICFG-PEDES~\cite{ding2021semantically}, and RSTPReid~\cite{zhu2021dssl}) and Video Anomaly Detection datasets (such as ShanghaiTech~\cite{luo2017revisit} and UBnormal~\cite{acsintoae2022ubnormal}) in Table \ref{tab:comparison}, focusing on image count, text annotation quantity and length, data sources, and annotation types. PAB features the following characteristics:
% \begin{itemize}[leftmargin=*]
% \item 
{\bf (1) A Large Number of Anomalous Behaviors:} Unlike general pedestrian datasets that focus on the viewpoint and occlusion of pedestrian images, PAB emphasizes providing a large number of images and textual descriptions of pedestrian anomaly. This complements ordinary Person Re-Identification tasks and presents new challenges.
% \item 
{\bf (2) High-Fidelity Images}: Compared to pedestrian images from surveillance cameras, which often suffer from poor lighting, blurry textures, 
% and face distortions or color mismatches, 
the images in PAB are of higher quality due to the generation method we adopted. The synthesized images are reasonable, realistic, and aesthetically pleasing.
% \item 
{\bf (3) Specific Textual Descriptions}: Relative to existing cross-modal pedestrian datasets, the texts in PAB are longer and provide more detailed information, including the appearance, clothing, actions, and background context of individuals. For pedestrian anomaly retrieval, information about the appearance and actions of individuals is crucial.
% \item 
{\bf (4) Diversity}: PAB encompasses a wide range of images that vary in terms of appearance, posture, viewpoint, background, and occlusion. Additionally, the text generation approach ensures that the image captions in PAB are sufficiently diverse, too.
% \item 
{\bf (5) Large-Scale Image-Text Pairs}: PAB comprises over one million image-text pairs, supporting deep cross-modal models in learning better uni-modal features and more inter-modal associations from the data.
% \item 
{\bf (6) Rich Annotations}: Each image-text pair in PAB is annotated with corresponding action, anomaly, and scene categories, providing additional data information besides person appearance.
% \item 
{\bf (7) Less Privacy Concerns}: Apart from a small amount of data in the test set sourced from existing public datasets, the vast majority of the data in PAB is synthesized, reducing ethical and legal issues. 
% \end{itemize}

%% file: sec/4_method.tex
\section{Method}
As shown in Fig.~\ref{fig:framework}, we introduce Cross-Modal Pose-aware (CMP) framework, which includes a pose-aware image encoder, a text encoder, and a cross encoder. During training, we use our Identity-based Hard Negative Mining (IHNM) strategy to create challenging negative pairs. These pairs then undergo feature extraction through the respective encoders, followed by contrastive learning. The features are then processed by the cross encoder for multimodal encoding and pass through the Image-Text Matching (ITM) and Mask Language Modeling (MLM) heads. The final step involves calculating the IHNM-enhanced ITM loss and MLM loss. Hereinafter, we detail the three core components:  the pose-aware image encoder, the identity-based hard negative mining strategy, and the cross-modal modeling module.

\subsection{Pose-aware Image Encoder (PE)}
Considering that normal and anomaly actions notably differ in human pose, we devise a pose-aware image encoder that explicitly incorporates human pose to enhance behavior comprehension. 
For a given image $I$, a human key point detector~\cite{cao2017realtime} is utilized to obtain the pose map $P$. 
Given $P$, we extract pose-aware feature via the Pose Conv module, and then feed the feature into the image encoder to obtain $f_P$, while $I$ is directly input into the image encoder to extract image embedding $f_I$. The Pose Conv module comprises convolutional layers, batch normalization, and ReLU activation functions to align the input domain.

\noindent {\bf Image Encoder.}
Without loss of generality, we deploy Swin Transformer (Swin-B)~\cite{liu2021swin} as the Image Encoder. The input image is initially divided into non-overlapping patches, which are then linearly embedded. These embedded patches are subsequently processed by transformer blocks, comprising Multi-Head Self-Attention and Feed Forward modules, to generate patch embeddings. Each patch embedding encapsulates the information of its corresponding patch. To aggregate the information from all patches, we compute the average of their features, referred to as the \texttt{[CLS]} embedding, and prepend it to the sequence.

\noindent {\bf Pose-aware Cross-Attention.}
Pose representation $f_P$, after being regularized by Layer Norm, is integrated into the image representation through a multi-head cross-attention module. 
The output ($f_{CA}$) of the cross-attention can be defined as: $f_{CA} = Softmax(\frac{{qk}^T}{\sqrt{d}}){v}$, where ${q} =W_q f_P, {k}=W_k f_I, {v}=W_v f_I$.
$q,k,v$ are the query, key, and values matrices of the attention operation respectively, $d$ is the dimension of $k$, and $W_q, W_k, W_v$ are the weights of the linear layers. 
Then, we combine the output of CA with the image embedding as the pose-aware representation $f_V = f_I + f_{CA}$.

% The output ($f_{CA}$) of the cross-attention can be defined by the following equation:
% \begin{align}
%   f_{CA} &= Softmax(\frac{{\bf qk}^T}{\sqrt{d}}){\bf v},
%   \label{eq:CA} \\
%   s.t. \quad {\bf q} &=W_q f_P, \quad {\bf k}=W_k f_I, \quad {\bf v}=W_v f_I.
%   \label{eq:qkv}
% \end{align}

\subsection{Identity-based Hard Negative Mining (IHNM)}
The key to solving pedestrian anomaly research is effectively establishing the relationship between pedestrian descriptions (with normal or anomaly action) and images. Ideally, for each text, the model should be able to search for an image that matches the appearance, actions, and background described, particularly the action description. Training such a cross-modal search model requires large-scale matching image-text pairs, negative pairs, and hard negative pairs, which are crucial for enabling the model to learn more discriminative features. Based on the construction process of PAB, we propose an Identity-based hard negative pair mining (IHNM) method that provides corresponding hard negative samples for each image and text in the training set. The sampling strategy is illustrated in \Cref{fig:framework} (c). Consider a training pair $(I, T)$, where $I$ is generated from anomaly/normal caption $C\in \{C_a, C_{a+}, C_n\}$, and $T$ is its re-captioned text. As the normal/anomaly counterpart $\tilde{C}$ of $C$ (both describing the same person) is known according to the OOPS! dataset, we can pick $\tilde{I}$ generated from $\tilde{C}$, and its re-captioned text $\tilde{T}$ can also be chosen accordingly. 
As shown in the \Cref{fig:framework} (c), $C$ and $\tilde{C}$ describe the anomalous and normal behaviors of the same pedestrian, respectively. Consequently, the generated images have similar pedestrian appearances and backgrounds but different actions. Intuitively, the image $I$ and the text $\tilde{T}$ form an ID-based hard negative pair. 
The establishment of hard negative pairs with similar appearances and backgrounds but different actions is critical for the model to learn features that are more discriminative of actions.

\begin{table*}
\vspace{-.2in}
  \centering
  \footnotesize
  \resizebox{0.98\linewidth}{!}{
  \begin{tabular}{@{}l|cc|cc|cc|cc|cc|cc|cc|cc|cc|cc|cc@{}}
    \toprule
    \multirow{2}{*}{Method}  &  \multicolumn{2}{c|}{Normal}  &  \multicolumn{2}{c|}{Wind}  &  \multicolumn{2}{c|}{Rain}  &  \multicolumn{2}{c|}{Snow}  &  \multicolumn{2}{c|}{Rain + Snow}  &  \multicolumn{2}{c|}{Dark}  &  \multicolumn{2}{c|}{Dark + Wind}  &  \multicolumn{2}{c|}{Dark + Rain}  &  \multicolumn{2}{c|}{Dark + Snow}  &  \multicolumn{2}{c|}{Over-exposure}  &  \multicolumn{2}{c}{Mean $\uparrow$}  \\
    &  R@1  &  mAP  &  R@1  &  mAP  &  R@1  &  mAP  &  R@1  &  mAP  &  R@1  &  mAP  &  R@1  &  mAP  &  R@1  &  mAP  &  R@1  &  mAP  &  R@1  &  mAP  &  R@1  &  mAP  &  R@1  &  mAP  \\
    \midrule
    % \multicolumn{23}{c}{\emph{w/o} Distractors in Gallery} \\
    % \midrule
    Baseline   & 83.47 & 90.94 & 79.02 & 88.10 & 54.40 & 67.40 & 59.10 & 72.34 & 49.95 & 62.09 & 79.58 & 88.24 & 75.53 & 85.47 & 34.93 & 45.98 & 50.25 & 63.32 & 74.87 & 84.82 & 64.11 & 74.87 \\
    + PE       & 84.13 & 91.20 & 78.92 & 87.93 & 55.01 & 67.31 & 61.78 & 74.11 & 50.76 & 63.28 & 78.41 & 87.72 & 75.43 & 85.43 & 39.08 & 49.77 & 50.10 & 63.06 & 75.89 & 85.51 & 64.95 & 75.53 \\
    + IHNM     & 84.48 & 91.36 & 79.47 & 88.25 & 58.24 & 70.29 & 60.11 & 72.78 & 51.47 & 63.87 & 80.18 & 88.83 & 75.78 & 85.71 & {\bf 40.75} & {\bf 50.88} & 52.12 & 64.52 & {\bf 76.64} & {\bf 85.99} & 65.92 & 76.25 \\
    CMP (Ours) & {\bf 84.93} & {\bf 91.66} & {\bf 81.24} & {\bf 89.34} & {\bf 60.06} & {\bf 72.53} & {\bf 63.40} & {\bf 75.74} & {\bf 54.85} & {\bf 67.31} & {\bf 80.89} & {\bf 89.00} & {\bf 77.20} & {\bf 86.55} & 39.03 & 50.58 & {\bf 53.49} & {\bf 66.12} & 76.09 & 85.56 & {\bf 67.12} & {\bf 77.44} \\ 
    % \midrule
    % \multicolumn{23}{c}{\emph{w} Distractors in Gallery} \\
    % \midrule
    % Baseline   &  &  &  &  &  &  &  &  &  &  &  &  &  &  &  &  &  &  &  &  &  &  \\
    % + PE       &  &  &  &  &  &  &  &  &  &  &  &  &  &  &  &  &  &  &  &  &  &  \\
    % + IHNM     &  &  &  &  &  &  &  &  &  &  &  &  &  &  &  &  &  &  &  &  &  &  \\
    % CMP (Ours) & {\bf 77.25} & 86.44 &  &  &  &  &  &  &  &  &  &  &  &  &  &  &  &  &  &  &  &  \\
    \bottomrule
  \end{tabular}
  }
  \vspace{-.1in}
  \caption{Robust text-based person anomaly retrieval results on PAB under multi-weather setting.
  % with gallery images featuring 10 different environmental styles.
  % (\emph{w/o} and \emph{w} distractors in gallery)
  }
   \vspace{-.15in}
  \label{tab:multi}
\end{table*}

\begin{table}
  % \vspace{-.1in}
  \centering
  \footnotesize
  %\small
  \begin{tabular}{@{}l|c|cccc@{}}
    \toprule
    Method  &  \#Data  &  R@1  &  R@5  &  R@10  &  mAP  \\
    \midrule
    MRA~\cite{yang2025minimizing} & - & 9.91 & 23.66 & 31.45 & 17.15 \\
    RaSa~\cite{bai2023rasa} & - & 21.74 & 27.30 & 27.96 & 24.35 \\ 
    WoRA~\cite{sun2025data} & - & 22.25 & 45.91 & 53.54 & 33.39 \\
    APTM~\cite{yang2023towards} & - & 22.90 & 45.80 & 52.38 & 33.56 \\
    % APTM~\cite{yang2023towards} & - & 23.00 & 45.45 & 52.07 & 33.49 \\
    CAMeL~\cite{yu2025camel} & - & 24.47 & 50.00 & 58.75 & 36.75 \\
    IRRA~\cite{jiang2023cross} & - & 30.59 & 59.61 & 68.91 & 44.41 \\
    CLIP~\cite{radford2021learning} & - & 47.57 & 81.55 & 89.03 & 62.73 \\  
    X-VLM~\cite{xvlm} & - & 71.94 & 97.78 & 98.99 & 83.96 \\
    \midrule
    MRA~\cite{yang2025minimizing} & 0.1M & 70.53 & 94.69 & 97.47 & 81.59 \\
    APTM~\cite{yang2023towards} & 0.1M & 72.14 & 95.30 & 97.17 & 82.78 \\
    % APTM~\cite{yang2023towards} & 0.1M & 72.55 & 95.10 & 97.37 &  83.26\\
    CAMeL~\cite{yu2025camel} & 0.1M & 74.30 & 96.79 & 98.84 & 84.20 \\
    WoRA~\cite{sun2025data} & 0.1M & 74.47 & 96.82 & 98.48 & 84.60 \\
    IRRA~\cite{jiang2023cross} & 0.1M & 76.39 & 97.62 & 99.14 & 86.33 \\
    CLIP~\cite{radford2021learning} & 0.1M & 77.60 & 98.84 & {\bf 99.75} & 87.35 \\
    RaSa~\cite{bai2023rasa} & 0.1M & 80.79 & \underline{98.89} & 99.65 & 89.20 \\
    X-VLM~\cite{xvlm} & 0.1M & 81.95 & 98.84 & 99.19 & 89.86 \\
    \midrule
    CMP (Ours) & 0.1M & \underline{83.06} & \underline{98.89} & 99.49 & \underline{90.41} \\
    CMP (Ours) & 1M & {\bf 84.93} & {\bf 99.09} & {\bf 99.75} & {\bf 91.66} \\
    \bottomrule
  \end{tabular}
  \vspace{-.1in}
  \caption{{\bf Quantitative results of our proposed method and compared methods on the proposed dataset PAB.} The best result is indicated in \textbf{bold}, while the second best is \underline{underlined}. 
  }
  \vspace{-.1in}
  \label{tab:result}
\end{table}

\subsection{Cross-modal Modeling}
As shown in \Cref{fig:framework}, we optimize our proposed Cross-Modal Pose-aware model by three cross-modal modeling tasks, \ie, Image-Text Matching (ITM), Contrastive Learning, and Masked Language Modeling (MLM).
These tasks align texts and images integrated with the pose.

\noindent {\bf IHNM-Enhanced Image-Text Matching.} Image-text matching 
determines whether an (image, text) pair is matched or not. Typically, well-aligned pairs are considered positive samples. Rather than randomly selecting trivial negatives, we employ our IHNM strategy to identify hard negative pairs. These pairs differ only in the action with its positive counterpart, requiring the model to discern subtle action differences, a capability crucial for the task of person anomaly search. 
Given an image-text pair, their multi-modal feature is encoded by feeding their representation into a cross encoder.
Specifically, we use the last six layers of BERT as the cross encoder. The cross encoder takes the text embeddings as input and fuses the image embeddings in the cross-attention module at each layer. Similar to the multi-head cross-attention in PE, the text embeddings serve as queries (q), while the image embeddings act as keys (k) and values (v).
The \texttt{[CLS]} embedding of features output by the cross encoder, is projected into 2-dimensional space via an ITM head (\ie, an MLP), yielding the predicted image-text matching probability $\hat{p}(I,T)$. The Image-Text Matching loss is defined as:
\begin{equation}
\label{l:itm}
\begin{split}
\mathcal{L}_{\text{itm}} &= - \mathbb{E}[p(I, T)\log \hat{p}(I, T)\\
&+ (1-p(I, T))\log (1-\hat{p}(I, T))],
\end{split}
\end{equation}
where $p$ denotes the ground-truth label. $p(I,T)= 1$ for positive pairs, $0$ for negative ones.

\noindent {\bf Image-Text Contrastive Learning.} This constrain targets to align image-text pair via contrastive learning~\cite{yang2023towards,bai2023rasa,jiang2023cross}.
Given the pose-aware image feature 
$f_V$, we split it to $\{f_{cls}, f_{pat}\}$, where $cls$ and $pat$ are the \texttt{CLS} embedding and patch embeddings, respectively. To obtain a comprehensive global image representation, we consider both the patch embedding and the \texttt{CLS} embedding: $f_v = FC([\text{AVG}(f_{pat}), f_{cls}])$,
% \begin{equation}
% \label{avgfc}
%     f_v = FC([\text{AVG}(f_{pat}), f_{cls}]),
% \end{equation}
where $\text{AVG}(f_{pat})$ is the average feature of tokens in $f_{pat}$.
Following previous practice~\cite{yang2023towards,bai2023rasa}, we deploy the initial six transformer layers of BERT~\cite{devlin2019bert} for text encoding. The text $T$ is tokenized and prefixed with a single \texttt{[CLS]} token before being input into the encoder. 
This process yields a text embedding $f_T$. Mirroring the image encoding, we apply the same procedure to $f_T$, combining text and class embeddings 
% as per Eq.~\ref{avgfc}, 
to obtain the global text representation $f_t$.
The image-to-text similarity within the batch is defined as follows:
\begin{equation}
\label{eq:i2t}
S_{\text{I2T}} = \frac{\exp(s(f_v, f_t)/\tau)}{\sum_{j=1}^{N}\exp(s(f_v, f_t^j)/\tau)},
\end{equation}
where $s(\cdot, \cdot)$ is the cosine similarity, $\tau$ is a temperature parameter.
Similarly, the text-to-image similarity is $S_{\text{T2I}}$.
Finally, the contrastive learning loss is presented below:
\begin{equation}
\label{l:cl}
\begin{split}
\mathcal{L}_{\text{cl}} = -\frac{1}{2} \mathbb{E}[\log S_{\text{I2T}} + \log S_{\text{T2I}}].
\end{split}
\end{equation}

\noindent {\bf Mask Language Modeling (MLM)} predicts the masked words in the text based on the matched image.
We randomly enable the masking strategy\footnote{The masking strategy is: $10 \%$ of the tokens are replaced with random tokens, $10 \%$ remain unchanged, and $80 \%$ are replaced with \texttt{[MASK]}.} with a probability of $25 \%$. 
The person image $I$ and the corresponding masked text $\hat{T}$ pair are then fed into the cross encoder, followed by an MLM head (an MLP with softmax) to predict the masked tokens.
We minimize the cross-entropy loss:
\begin{equation}
\label{l:mlm}
\mathcal{L}_{\text{mlm}} = -\mathbb{E}[p_{\text{mask}}(I, \hat{T}) log(\hat{p}_{\text{mask}}(I, \hat{T}))],
\end{equation}
where $\hat{p}_{\text{mask}}$ is the predicted likelihood of the masked token $t$ in $\hat{T}$, $p_{\text{mask}}$ is the ground-truth one-hot vector. 
% Finally, 
The overall objective %of our proposed Cross-Modal Pose-aware framework 
can be formulated as: 
$\mathcal{L} = \mathcal{L}_{\text{cl}} + \mathcal{L}_{\text{itm}} +\mathcal{L}_{\text{mlm}}.$
% \begin{equation}
% \label{l:all}
% \mathcal{L} = \mathcal{L}_{\text{cl}} + \mathcal{L}_{\text{itm}} +\mathcal{L}_{\text{mlm}}.
% \end{equation}

%% file: sec/5_experiment.tex
\section{Experiment}

\noindent {\bf Dataset and Evaluation Metrics.}
The constructed PAB serves as the evaluation benchmark. To evaluate the performance of text-based person anomaly search, we adopt recall rates R@K and mAP. 
Given a query text, all test images are ranked according to their matching probability with the query. If the image that perfectly matches the content of the text description (appearance, actions, background, \etc) is among the top k images, the search is considered successful.
Mean Average Precision (mAP) refers to the average area under the precision-recall curve across all queries. 
% We report R@1, R@5, R@10, and mAP. 
Higher recall rates and mAP indicates better results.

\noindent {\bf Implementation Details.}
We train the Cross-Modal Pose-aware (CMP) model for 30 epochs with mini-batch size of $22$. 
We adopt AdamW ~\cite{loshchilov2018decoupled} optimizer with a weight decay of $0.01$. 
The learning rate linearly decreases from $1\times10^{-4}$ to $1\times10^{-5}$. 
CMP has $230.8 M$ trainable parameters, with $86.8 M$, $66.4 M$, and $59.1 M$ parameters for the image, text, and cross encoders, respectively.
The encoder weights are initialized by X-VLM~\cite{xvlm}. 

\noindent {\bf Quantitative Results.}
In \Cref{tab:result}, we compare our method with a wide range of possible solutions, including two state-of-the-art vision-language pre-training models (CLIP~\cite{radford2021learning} and X-VLM~\cite{xvlm}) and six state-of-the-art text-based person search methods.
% (MRA~\cite{yang2025minimizing}, CAMeL~\cite{yu2025camel}, WoRA~\cite{sun2025data}, APTM~\cite{yang2023towards}, IRRA~\cite{jiang2023cross}, and RaSa~\cite{bai2023rasa}) and .
% Firstly, the zero-shot transfer results in PAB indicate the challenge of our proposed data set.
% Firstly, CLIP~\cite{radford2021learning} achieves the highest zero-shot results, with $47.57 \%$ R@1, $81.55 \%$ R@5, $89.03 \%$ R@10, and $62.73 \%$ mAP, indicating that PAB presents a significant challenge.
Firstly, X-VLM~\cite{xvlm} achieves the fairly-good zero-shot results, with $71.94 \%$ R@1, and $83.96 \%$ mAP, indicating that the fine-grained action understanding remains a significant challenge.
After training 30 epochs on the same $0.1M$ image-text pairs from PAB, X-VLM~\cite{xvlm} achieves $81.95 \%$ R@1, significantly higher than the other four methods. Therefore, we adopt the X-VLM~\cite{xvlm} model to form a strong baseline.
Compared to Baseline, our Cross-Modal Pose-aware (CMP) method employs Pose-aware Image Encoder (PE) and Identity-based Hard Negative Mining (IHNM) and achieves an R@1 of $83.06 \%$, representing a $+1.11 \%$ increase over the Baseline model.
If all PAB data ($1M$) is adopted to train CMP, the recall@1 is $84.93 \%$, improving $+1.87 \%$ compared to $0.1M$ training images.
Similar trends are observed in other evaluation metrics.
% , \ie, R@5, R@10, and mAP. %These results verify the effectiveness of CMP. 
% In the supplementary file, we provide zero-shot results on PAB to show the challenge of our proposed dataset and evaluation on the real-world OOD test set.

\begin{table}[t]
  \centering
  \footnotesize
  %\small
  \begin{tabular}{@{}l|cccc@{}}
    \toprule
    Method  &  R@1  &  R@5  &  R@10  &  mAP  \\
    \midrule
    APTM~\cite{yang2023towards} & 27.86 & 40.41 & 46.77 & 22.61 \\
    % APTM~\cite{yang2023towards} & 30.00 & 44.38 & 51.39 & 24.49 \\
    IRRA~\cite{jiang2023cross} & 40.28 & 57.24 & 65.98 & 33.53 \\
    CLIP~\cite{radford2021learning} & 51.60 & 68.31 & 76.43 & 43.05 \\
    X-VLM~\cite{xvlm} & 52.33 & 66.73 & 72.54 & 40.87 \\
    RaSa~\cite{bai2023rasa} & \underline{54.12} & 70.32 & 75.96 & 39.71 \\
    \midrule
    CMP & \underline{54.12} & \underline{71.07} & \underline{77.90} & \underline{43.13} \\
    CMP (1M) & \textbf{55.23} & \textbf{71.67} & \textbf{77.99} & \textbf{44.35} \\
    \bottomrule
  \end{tabular}
  \vspace{-.1in}
  \caption{Comparisons with existing methods in OOD setting. The unseen test set UCC is extracted from the UCF-Crime~\cite{sultani2018real} dataset. %The best result is indicated in \textbf{bold}, while the second best is \underline{underlined}.
  }
  \vspace{-.1in}
  \label{tab:ood}
\end{table}

\begin{table}
  % \vspace{-.1in}
  \centering
  \footnotesize
  \resizebox{0.9\linewidth}{!}{
  \begin{tabular}{@{}l|cccc@{}}
    \toprule
    Evaluation setting & R@1 &  R@5 &  R@10 &  mAP \\
    \midrule
    %\shline
    Identity Search (Traditional) & 94.34 & 99.39 & 99.85 & 88.01 \\
    Behavior Search (Ours) & 84.93 & 99.09 & 99.75 & 91.66 \\
    \bottomrule
  \end{tabular}
  }
  \vspace{-.1in}
  \caption{Comparison of different evaluation settings with CMP.}
  \vspace{-.15in}
  \label{tab:setting}
\end{table}

\begin{figure*}[t]
  \centering
  \vspace{-.2in}
   \includegraphics[width=0.98\linewidth]{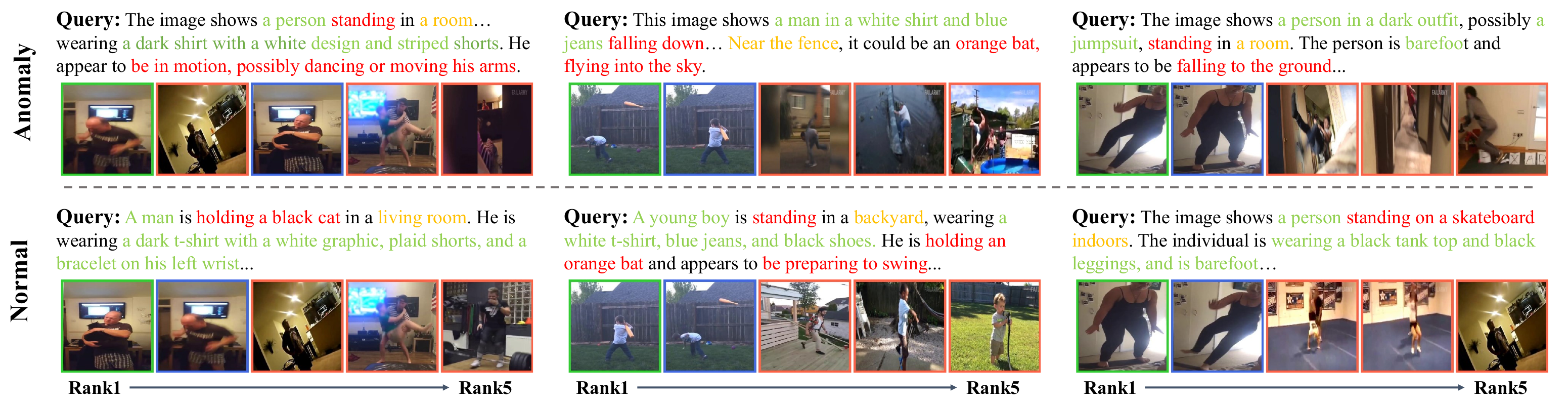}
   \vspace{-.1in}
   \caption{{\bf Qualitative Results.} 
   Top-5 anomaly search results for text queries: anomaly actions (top) and normal actions (bottom). Green rectangles indicate correct matches, red for mismatches, and blue for ID matches with behavior mismatches. Query parts (appearance, action, background) are highlighted in green, red, and orange, respectively.
   }%Examples of top-5 person anomaly search results with text queries for anomaly actions (top) and normal actions (bottom). Matched images are marked by green rectangles, mismatched images are  in red, and blue boxes indicate cases where the ID matches but the behavior does not. The parts of the queries that describe appearance, action, and background are highlighted in green, red, and orange.}
   \label{fig:result}
   \vspace{-.15in}
\end{figure*}

\begin{table}
  \centering
  \footnotesize
  \resizebox{0.9\linewidth}{!}{
  \begin{tabular}{@{}l|cc|cccc@{}}
    \toprule
    Method  &  PE  &  IHNM  &  R@1  &  R@5  &  R@10  &  mAP  \\
    \midrule
    Baseline &  &  & 83.47 & 99.04 & {\bf 99.80} & 90.94 \\
    M1 & $\checkmark$ &  & 84.13 & {\bf 99.14} & 99.65 & 91.20 \\
    M2 &  & $\checkmark$ & 84.48 & 98.94 & 99.60 & 91.36 \\
    CMP (Ours) & $\checkmark$ & $\checkmark$ & {\bf 84.93} & 99.09 & 99.75 & {\bf 91.66} \\
    \bottomrule
  \end{tabular}
  }
  \vspace{-.1in}
  \caption{Ablation studies on the key component of our proposed method, \ie, Pose-aware Image Encoder (PE) and Identity-based Hard Negative Mining (IHNM).}
   \vspace{-.15in}
  \label{tab:abl}
\end{table}

% \noindent {\bf Zero-Shot Transfer Setting.}
% \synote{In \Cref{tab:zero}, we present several zero-shot transfer results for text-based person anomaly search on the PAB dataset. Three state-of-the-art text-based person search models (APTM~\cite{yang2023towards}, IRRA~\cite{jiang2023cross}, and RaSa~\cite{bai2023rasa}), along with two leading vision-language pre-training methods (CLIP~\cite{radford2021learning} and X-VLM~\cite{xvlm}), are utilized to evaluate the quality of our proposed dataset. For this evaluation, these five models are directly applied to assess the performance of text-based person anomaly search on PAB without any training or fine-tuning on this dataset. Notably, CLIP~\cite{radford2021learning} achieves the highest results, with $47.57 \%$ R@1, $81.55 \%$ R@5, $89.03 \%$ R@10, and $62.73 \%$ mAP, indicating that PAB presents a significant challenge.}

\noindent {\bf Multi-weather Setting.}
% We further evaluate our CMP framework with and without the key component, \ie, Pose-aware Image Encoder (PE) and Identity-based Hard Negative Mining (IHNM), against 10 different conditional environments, \eg, wind/rain/snow (following \cite{wang2024multiple}). The results are shown in Table \ref{tab:multi}. CMP proves pivotal for robustness, showing consistent gains across all scenarios.
Following MuSe-Net~\cite{wang2024multiple}, we introduce a multi-weather test setting to simulate the round-the-clock (24/7) smart city scenario. Specifically, we evaluate the CMP framework with and without the key component, \ie, Pose-aware Image Encoder (PE) and Identity-based Hard Negative Mining (IHNM), under 10 distinct environmental conditions, \eg, wind/rain/snow. As shown in Table \ref{tab:multi}, CMP proves pivotal for robustness, showing consistent gains across all scenarios.

\noindent {\bf OOD Setting.}
To evaluate the scalability of the CMP model, we extract a new real-world test set from UCF-Crime~\cite{sultani2018real} for an Out-of-Distribution (OOD) test, and we call it UCC. Specifically, we select keyframes from $13$ types of abnormal and normal videos in UCF-Crime and leverage Qwen2-VL~\cite{bai2023qwen} to generate text descriptions for these video frames. It results $5,320$ image-text pairs as the OOD test set. As shown in \Cref{tab:ood}, our model CMP achieves superior performance, $54.12 \%$ R@1 and $43.13 \%$ mAP, compared to other state-of-the-art methods (also trained on $0.1M$ PAB data). After training with complete $1M$ PAB training data, CMP further improves by $1.11 \%$ in R@1. These experimental results show that our model exhibits competitive adaptability when facing unseen datasets. %, validating its robustness.

% We adopt the X-VLM model~\cite{xvlm}, which is the state-of-the-art method for text-to-image retrieval, to form a strong baseline. 
% After 30 epochs of training on the same $0.1M$ PAB data, our method outperforms RaSa~\cite{bai2023rasa} by $23.46 \%$ R@1 and $19.90 \%$ mAP.
% The Baseline is trained on $0.1M$ image-text pairs from PAB and achieves $69.92 \%$ R@1, significantly higher than the Zero-Shot performance of $9.40\%$. 
% This result indicates that the synthetic training data of PAB facilitates fine-grained behavior retrieval on real-world test sets.
% Compared to Baseline, our method employs Identity-based Hard Negative Mining (IHNM) and Pose-aware Image Encoder (PE). 
% After being trained on the same $0.1M$ data, our method achieves an R@1 of $72.80 \%$, representing a $2.88 \%$ improvement over the Baseline model. 

\noindent {\bf Behavior Search vs. Identity Search.} Text-based person anomaly search poses significantly greater challenges than traditional identity-based person search, as it requires a fine-grained understanding of both appearance and behavior. As shown in \Cref{tab:setting}, while traditional methods treat all images with the same ID as correct matches, our anomaly search task demands precise localization of specific behaviors. This distinction is evident in our experimental results: our model achieves 94.34\% R@1 on traditional text-based person search, outperforming its performance on anomaly search by 9.41\%. The performance gap stems from the fundamental difference between the two tasks: identity search primarily relies on appearance, whereas anomaly retrieval necessitates distinguishing subtle behavioral patterns. Such capability is crucial for real-world applications, including security event tracing, emergency response, and other scenarios where requires precise behavior localization.

%Text-based person anomaly search demands a finer-grained understanding of both the pedestrian's appearance and behavior, thus posing more challenges. \Cref{tab:setting} compares the traditional identity match (the images with the same ID are all treated as the correct matches) and the focused person anomaly search.Our model achieves $94.34 \%$ R@1 on the task of traditional Text-based Person Search, which is $9.41 \%$ higher than our proposed anomaly search. This is because, compared to pedestrian identity search, which mainly requires appearance matching. Anomaly retrieval is a more challenging task that demands the model to distinguish fine-grained behaviors. This task characteristic is critical for security event tracing and locating, emergency response, and many other security applications.

%We further explore the application of our method to Text-based Person Search. 
%Specifically, during inference, we consider all retrieved images of the same ID as correct matches (Identity Match), \ie, the images marked with green and blue boxes in \Cref{fig:result}. 
%The experimental results are shown in \Cref{tab:setting}, where 

\noindent {\bf Qualitative Results.}
We qualitatively evaluate our method on the task of Text-based Person Anomaly Search. \Cref{fig:result} shows six fine-grained pedestrian behavior retrieval results via the trained Cross-Modal Pose-aware (CMP) model, with queries for anomalous behavior (top) and normal behavior (bottom). 
For each query, we display the top five retrieved images. Correct retrieval results are marked with green boxes. Blue and red boxes indicate incorrect matches. Note that blue boxes denote images that belong to the same ID as the query but do not match the action. 
To further illustrate the retrieval effects, we highlight the parts of the text queries that describe appearance in green, action in red, and the background in orange. 
It can be observed that, in addition to appearance and background information, our model can effectively distinguish fine-grained action information. Even the incorrectly matched images displayed in \Cref{fig:result} still show reasonable relevance to the query sentences.

%\subsection{Ablation Studies and Further Discussion}
\noindent {\bf Effect of the Key Components.}
We conduct ablation studies on the key components of the Cross-Modal Pose-aware (CMP) model.
As shown in \Cref{tab:abl}, for fairness, all variants are trained for $30$ epochs on $1M$ image-text pairs. 
First, we evaluate the effectiveness of Pose-aware Image Encoder (PE) by comparing the Baseline and M1 models. It indicates that Pose-aware Image Encoder (PE) is a critical component of CMP. 
Furthermore, M2 incorporating Identity-based Hard Negative Mining (IHNM) obtains $1.01 \%$ R@1 improvement compared to the Baseline. The results show that adding IHNM improves the ability in discriminating the fine-grained behavior. %retrieval performance, indicating the necessity of identity-based hard negative samples.
Our CMP method equipped with both PE and IHNM arrives at the best performance. %significant improvements over the Baseline, M1, and M2 models
%Specifically, it achieves an R@1 of $84.93\%$, representing a substantial enhancement in retrieval performance.
%We observe that adopting only $10 \%$ data can achieve $98 \%$ performance.

\begin{figure}[t]
  \centering
   \includegraphics[width=0.85\linewidth]{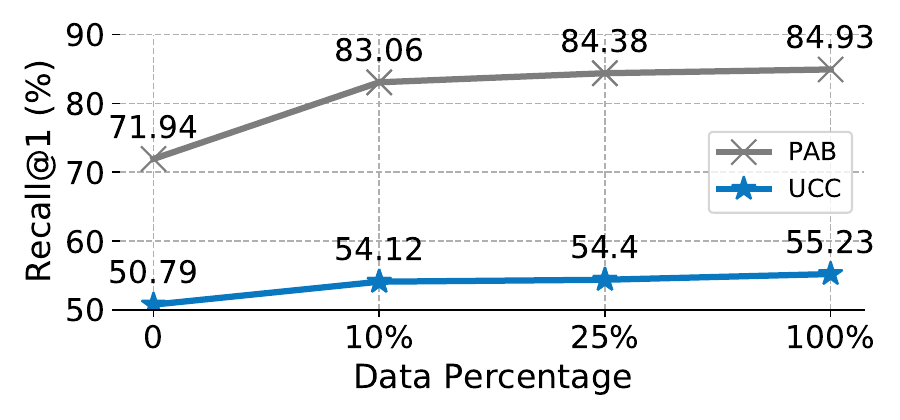}
   \vspace{-.15in}
   \caption{{\bf Ablation studies on synthetic training data scale.} We gradually increase PAB training data from 0\% to 100\% .% to train and then report the recall rate on the real-world test set.
   }
   \label{fig:scale}
   \vspace{-.2in}
\end{figure}

\noindent{\bf Effect of Training Data.} We further study the impact of training data scale on model performance in Fig.~\ref{fig:scale}. Considering the hard negative sampling, our method with 10\% training data has already achieves competitive recall rate.

%% file: sec/6_conclusion.tex
\section{Conclusion}
% We introduce Text-based Person Anomaly Search, a new task addressing the limitation of traditional retrieval in identifying anomalies. To support this, we construct the Pedestrian Anomaly Behavior (PAB) benchmark, combining real-world test videos with synthetic training data. PAB provides a large-scale training set with fine-grained annotations and a high anomaly ratio, facilitating the model learning. Our Cross-Modal Pose-aware (CMP) framework integrates pose patterns and hard negative mining to distinguish between normal and anomalous actions. Extensive experiments on PAB and UCC verify the effectiveness of our Cross-Modal Pose (CMP) model in retrieving both anomalous and normal behaviors in real-world scenarios.
% PAB provides fine-grained annotations and a high anomaly ratio.
We propose Text-based Person Anomaly Search, a new task that overcomes the limitation of traditional retrieval in anomaly identification. To support this, we introduce the large-scale Pedestrian Anomaly Behavior (PAB) benchmark, combining synthetic training data with real-world test frames. 
Our Cross-Modal Pose-aware (CMP) framework leverages pose patterns and hard negative mining to distinguish normal/anomalous actions. Extensive experiments on PAB and UCC confirm the effectiveness of CMP in retrieving both anomalies and normal behaviors in real-world scenarios and multi-weather settings.

\noindent\textbf{Acknowledgments}
% This work is supported by the National Key Research and Development Project, China (No. 2019YFB2102501).
We acknowledge supports from the Macau Science and Technology Development Fund FDCT/0043/2025/RIA1 and the Nanjing Municipal Science and Technology Bureau 202401035. This work is also supported in part by the Fundamental Research Funds for the Central Universities with No. JZ2024HGTB0261 and the NSFC project under grant No. 62302140.

%% file: sec/X_suppl.tex
\clearpage
\setcounter{table}{6}
\setcounter{figure}{5}
% \setcounter{section}{6}
% \maketitle

% \pagestyle{plain}
\twocolumn[{
\renewcommand\twocolumn[1][]{#1}
\maketitlesupplementary
\begin{center}
\setlength{\abovecaptionskip}{8pt} 
\setlength{\belowcaptionskip}{0pt} 
\centering
\vspace{-.05in}
\includegraphics[height=2.4in, width=6.8in]{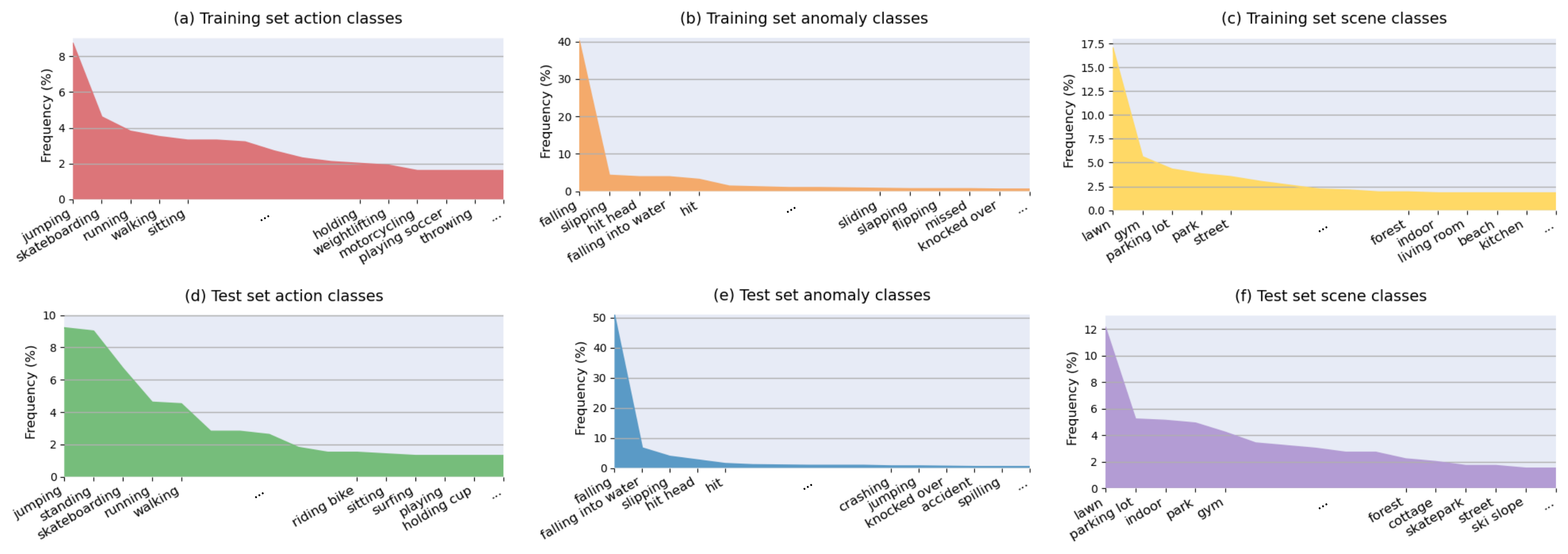}
\vspace{-.05in}
\captionof{figure}
{\textbf{Dataset Statistics}. 
An overview of the attribute annotations, including the distribution of categories across the training and test sets. Specifically, it covers {normal} action categories {(a, d)}, anomaly categories {(b, e)}, and scene categories {(c, f)}. Due to the natural long-tail distribution of the data {and the space limitation}, we present the top 15 most common classes for each category to ensure clarity. (Best viewed when zooming in.)
}
\label{fig:statistics}
\end{center}}]

\begin{table*}[t]
  \centering
  \footnotesize
  % \small
  \begin{tabular}{@{}l|cccccccc@{}}
    \toprule
    \multirow{1}{*}{Datasets} & \multirow{1}{*}{Modality} & \multirow{1}{*}{\#Frames} & \multirow{1}{*}{\#Scenes} & \#Anomaly Types & Anomaly:Normal & \#Avg Words & Open Set & Data Source \\
    % &  &  &  & Types &  & Set &  \\
    \midrule
    UCSD Ped2~\cite{li2013anomaly} & Video & 4,560 & 1 & 5 Classes & 1:2  & - & $\checkmark$ & Collection \\
    UMN~\cite{mehran2009abnormal} & Video & 7,741 & 3 & 1 Classes & 1:4  & - & $\checkmark$ & Collection \\
    UCSD Ped1~\cite{li2013anomaly} & Video & 14,000 & 1 & 5 Classes & 1:2  & - & $\checkmark$ & Collection \\   
    CUHK Avenue~\cite{lu2013abnormal} & Video & 30,652 & 1 & 5 Classes & 1:7 & - & $\checkmark$ & Collection \\
    Subway Exit~\cite{adam2008robust} & Video & 64,901 & 1 & 3 Classes & 1:13  & - & $\checkmark$ & Collection \\
    Subway Entrance~\cite{adam2008robust} & Video & 144,250 & 1 & 5 Classes & 1:11  & - & $\checkmark$ & Collection \\
    Street Scene~\cite{ramachandra2020street} & Video & 203,257 & 1 & 17 Classes & 1:4  & - & $\checkmark$ & Collection \\
    UBnormal~\cite{acsintoae2022ubnormal} & Video & 236,902 & 29 & 22 Classes & 2:3  & - & $\checkmark$ & Synthesis \\
    ShanghaiTech~\cite{luo2017revisit} & Video & 317,398 & 13 & 11 Classes & 1:18  & - & $\checkmark$ & Collection \\
    UCF-Crime~\cite{sultani2018real} & Video & {13,741,393} & Unlimited & 13 Classes & $\ll$1:1  & - & $\times$ & Collection \\
    UCA~\cite{yuan2024towards} & Video, Text & 13,741,393 & Unlimited & 13 Classes & $\ll$1:1 & 20.2 & $\times$ & Collection \\
    \midrule
    \multirow{2}{*}{{PAB} (Ours)} & \multirow{2}{*}{Image, Text} & \multirow{2}{*}{1,015,583} & \multirow{2}{*}{\bf 480} & \multirow{2}{*}{{\bf 1600} Classes} & \multirow{2}{*}{\bf 3:2} & \multirow{2}{*}{50.3} & \multirow{2}{*}{$\checkmark$} & Synthesis \& \\
    &  &  &  &  &  &  &  & Collection \\
    \bottomrule
  \end{tabular}
  % \vspace{-.1in}
  \caption{Comparison of the statistics of our PAB and other Video Anomaly Detection (VAD) datasets. 
  % ``Ano." and ``Nor." mean ``Anomaly" and ``Normal". 
  The statistics of previous datasets have been recorded in~\cite{acsintoae2022ubnormal}.
  }
  % \vspace{-.1in}
  \label{tab:video}
\end{table*}

\begin{figure*}[t]
  \centering
   \includegraphics[width=1\linewidth]{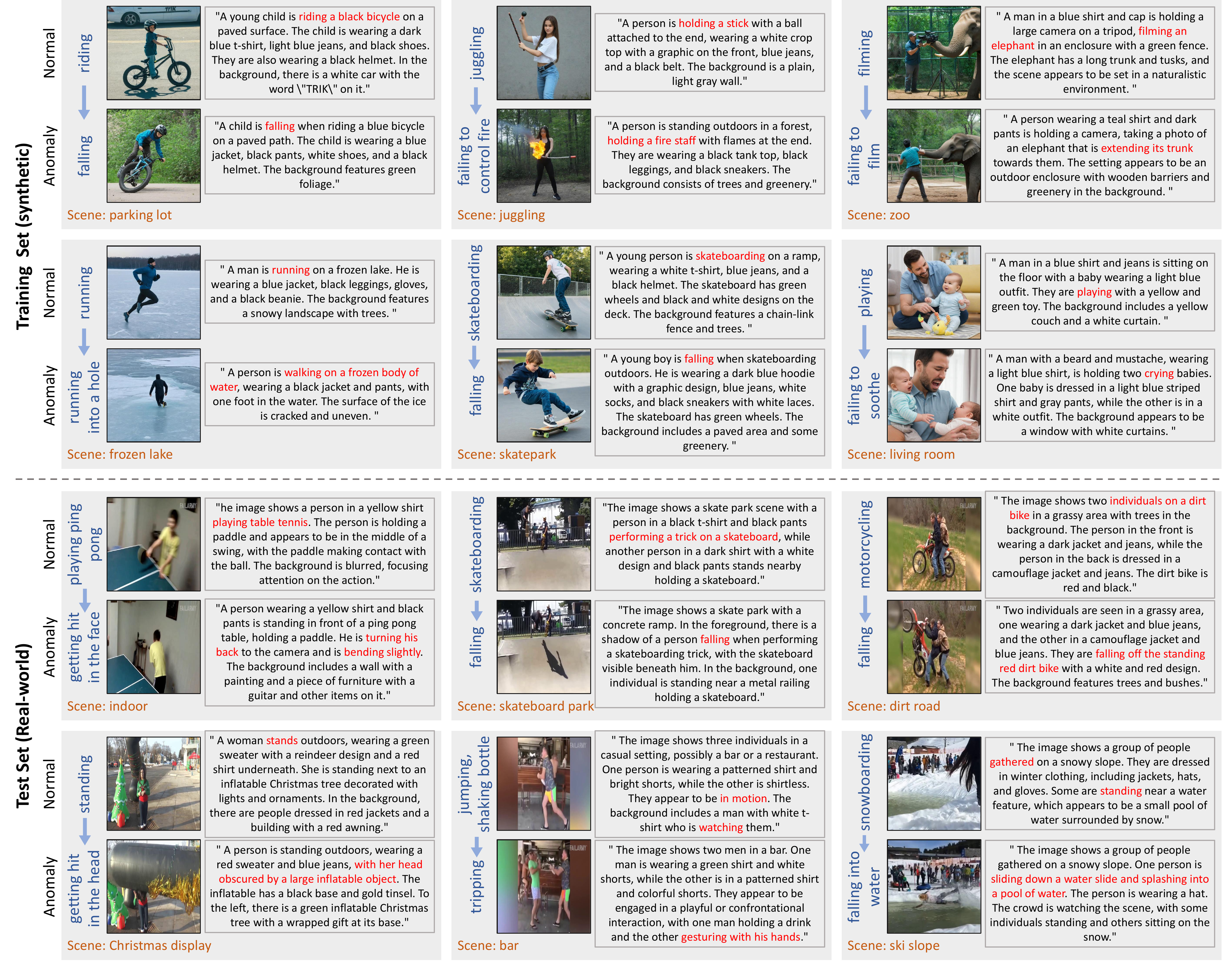}
   \caption{\textbf{Dataset Examples}. 12 training (synthetic) image-text pairs from the PAB dataset {are at the top}, while 12 test (real-world) image-text pairs {are at the bottom}. Half of the examples depict anomaly behaviors, while the other half show corresponding normal actions. Each pair is annotated with scene and action (or anomaly) classes. Minor errors may be present in the generated captions of the training set, whereas the captions in the test set have been refined by professionals. 
   (Best viewed on a computer screen with zoom.)}
   \label{fig:more}
\end{figure*}

% \section{Supplementary}
\section*{Appendix}

\section{More Benchmark Details}
\subsection{Attribute Annotation Details.}
During the attribute annotation process, we utilize {the widely-used} Qwen2-VL~\cite{bai2023qwen} to annotate each normal image-text pair with an action type and scene category, while each anomaly image-text pair is annotated with an anomalous behavior class and scene category.
% Consider a training pair $(I, T)$, where $I$ is generated from anomaly/normal caption $C\in \{C_a, C_{a+}, C_n\}$, and $T$ is its re-captioned text.
For a given image-text pair ($I, T$), if ($I, T$) is a training pair, $I$ is generated from anomaly/normal caption $C\in \{C_a, C_{a+}, C_n\}$, and $T$ is its re-captioned text.
If ($I, T$) is a test pair, $C\in \{C_a, C_n\}$ is the caption of corresponding source video and $T$ is the re-captioned text for $I$.
We leverage $I, C$ to design instructions and query the MLLM for attribute.
The specific {\bf Instructions} are as follows:
\begin{itemize}
    \item {\bf Instruction for Anomaly Behavior Class:} ``Below is the image caption of the image. In the image, someone fails to do something. Based on the caption and image, summarize the failure of the characters in the image using a single word or phrase, such as falling, losing balance, slipping, falling to the ground, falling into water, losing control, having accident, flipping, jumping, hitting head, \etc. Image caption: $C$."
    \item {\bf Instruction for Action Type:} ``Below is the image caption of the image. Based on the caption and image, summarize the behavior and action categories of the characters in the image using a single word or phrase, such as motorcycling, driving car, somersaulting, riding scooter, catching fish, staring at someone, dyeing eyebrows, trimming beard, peeling potatoes, square dancing, \etc. Image caption: $C$."
    \item {\bf Instruction for Scene Category:} ``Below is the image caption of the image. Based on the caption and image, summarize the scene or background of the characters in the image using a single word or phrase, such as playground, parking lot, ski slope, highway, lawn, outdoor church, cottage, indoor flea market, fabric store, hotel, \etc. Image caption: $C$."
\end{itemize}

\subsection{Attribute Statistics.}
Based on the three types of instructions, we automatically obtain action, anomaly, and scene attributes. As shown in \Cref{fig:statistics}, we present the distribution of the top 15 most common classes for each attribute in both the training and test sets. The attribute distributions in both sets are similar and naturally exhibit a long-tail distribution
For action types, the top five in the training set are jumping, skateboarding, running, walking, and sitting, while the top five in the test set are jumping, standing, skateboarding, running, and walking (\Cref{fig:statistics} (a) and (d)).
The most frequent anomalous behavior is falling, occurring with approximately $40 \%$ frequency in the training set (\Cref{fig:statistics} (b)) and $50 \%$ in the test set (\Cref{fig:statistics} (e)).
The scene distribution is primarily concentrated on the lawn, gym, and parking lot in both subsets, as shown in \Cref{fig:statistics} (c) and (f).

\subsection{Comparisons with More Video Anomaly Detection Datasets.}
{In \Cref{tab:video}, we compare our proposed PAB dataset with the most utilized Video Anomaly Detection (VAD) datasets. Eight metrics are reported: modality, number of frames/images, scenes, anomaly types, the proportion of anomaly versus normal, average number of words per sentence, open-set characteristics, and data source. Compared to other video datasets, PAB is distinguished as an image-text pair dataset and features a higher number of anomaly types from a broader range of event scenes. For all datasets, the ``Anomaly:Normal" ratio represents the proportion of anomaly video frames/images to normal video frames/images. 
While most VAD datasets are annotated solely with normal/abnormal labels or abnormal category labels, PAB provides detailed annotations including appearance descriptions, actions, and scene information. Most video datasets maintain open-set characteristics for anomaly detection. To ensure consistent open-set characteristics, we provide a real-world Out-of-Distribution (OOD) test set for PAB sourced from UCF-Crime~\cite{sultani2018real}. 
Notably, UBnormal~\cite{acsintoae2022ubnormal} is also a synthetic dataset, but unlike PAB, both its training and test sets consist entirely of synthesized data.}

\subsection{Visualizations.}
In Figure \ref{fig:more}, we present additional example image-text pairs from our proposed dataset, PAB. The figure includes 12 synthetic training image-text pairs {(top)} and 12 real-world test image-text pairs {(bottom)}. These pairs are divided into two categories: %half 
{one} depicts anomalous behaviors, while %the other half 
{the other} illustrates normal actions. Each image-text pair is meticulously annotated with specific scene and action (or anomaly) classifications to facilitate further precise learning and evaluation.
It is worth noting that while the training set sometimes contains some {noise} %minor inaccuracies 
in the generated captions, the test set captions have been professionally refined to ensure high-quality annotations. This provides a reliable benchmark for assessing model performance.

\begin{figure*}[t]
  \centering
   \includegraphics[width=1\linewidth]{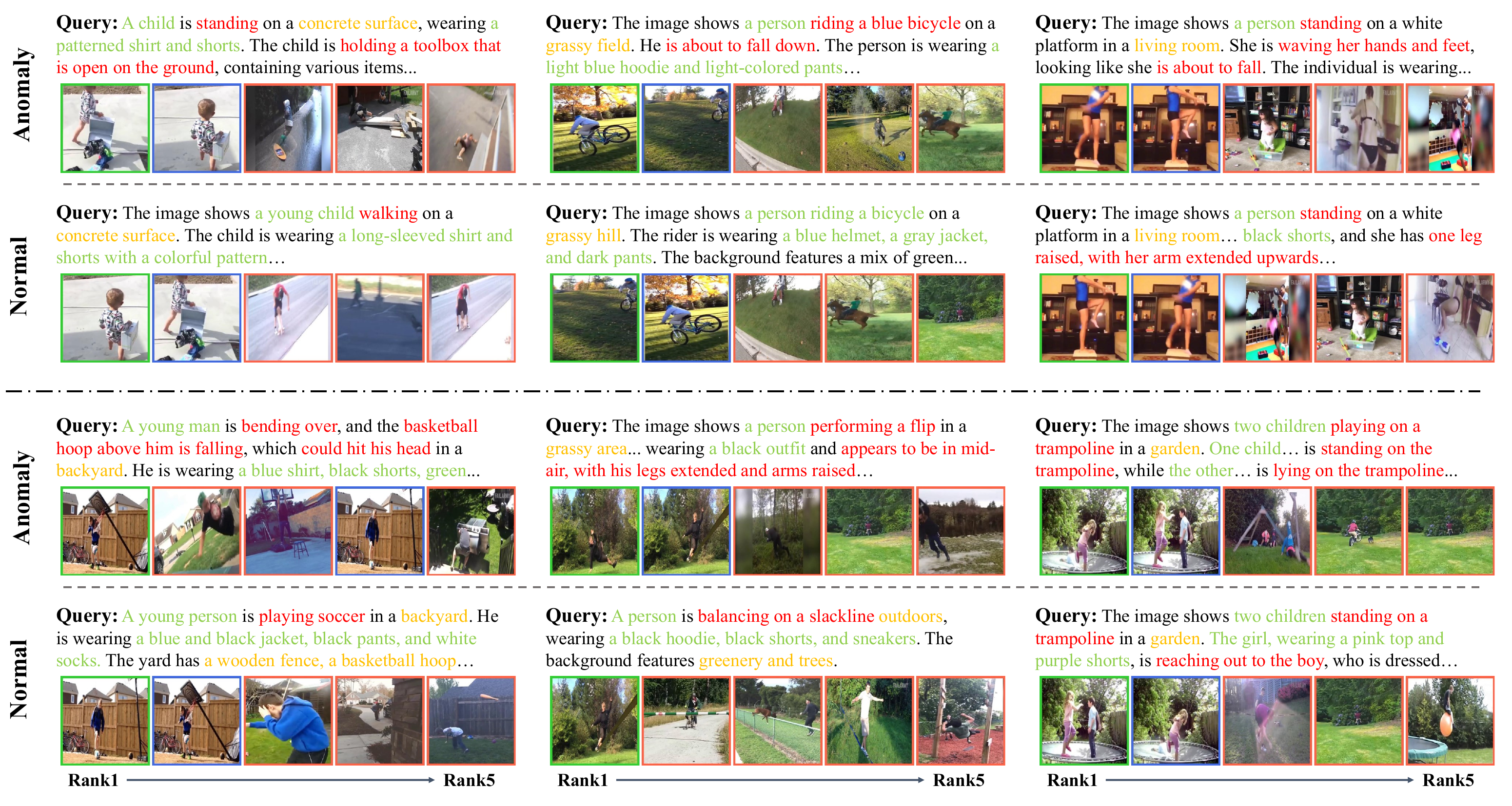}
   \caption{\textbf{More Qualitative Results.} 12 examples of top-5 person anomaly search results with text queries for anomaly actions and normal actions. Matched images are marked by green boxes, mismatched images are marked in red, and blue boxes indicate cases where the ID matches but the behavior does not. The parts of the queries that describe appearance, action, and background are highlighted in green, red, and orange.
   It is best viewed on a computer screen with Zoom.}
   \label{fig:result-more}
\end{figure*}

\section{Experiment Details and Further Experiments}
\subsection{Training Details.}
We train the Cross-Modal Pose-aware (CMP) model using PyTorch on four NVIDIA GeForce RTX 3090 GPUs. The first $500$ training iterations serve as a warm-up phase. Each image input is resized to $224 \times 224$ pixels, and the maximum text token length is set to $56$. For image augmentation, we apply techniques such as random horizontal flipping and random erasing. For text augmentation, we employ EDA~\cite{wei2019eda}. Training for 30 epochs on the full training set takes approximately 4 days and 4 hours.

\subsection{Inference Details.}
During inference, we first obtain the embeddings of all query texts and candidate images (integrated with pose) from the test set, then compute the text-to-image similarity. For each query, we select the top $128$ images with the highest similarity scores. These images are then re-ranked based on the matching probabilities predicted by the cross-modal encoder and the MLP head. The final ranking results constitute the search outcomes of the model.

\subsection{More Qualitative Result Examples.}
We present 12 additional text-based person anomaly search qualitative results of our method in Figure \ref{fig:result-more}. For each query (anomalous or normal), we display the top five retrieved images. True retrieval results are marked with green boxes, while blue and red boxes indicate incorrect matches. Blue boxes denote images that belong to the same ID as the query but do not match the action.
We highlight the parts of the text queries that describe appearance in green, action in red, and the background in orange. In addition to appearance and background information, our model can effectively distinguish fine-grained action information. Even the incorrectly matched images displayed in Figure \ref{fig:result-more} still show some relevance to the query sentences.